# Less is More: Genetic Optimisation of Nearest Neighbour Classifiers

Vitorino Ramos, Fernando Muge

*Centro de Valorização de Recursos Minerais, Instituto Superior Técnico*
*Technical University of Lisbon*
*Av. Rovisco Pais - 1096 Lisboa Codex, PORTUGAL*
*email:{vitorino.ramos,pcmuge}@alfa.ist.utl.pt*

**Abstract:** The present paper deals with optimisation of *Nearest Neighbour* rule Classifiers via *Genetic Algorithms*. The methodology consists on implement a Genetic Algorithm capable of search the input feature space used by the *NNR* classifier. Results show that is adequate to perform feature reduction and simultaneous improve the Recognition Rate. Some practical examples prove that is possible to Recognise Portuguese Granites in 100%, with only 3 morphological features (from an original set of 117 features), which is well suited for real time applications. Moreover, the present method represents a robust strategy to understand the proper nature of the images treated, and their discriminant features.

**Keywords:** *Feature Reduction, Genetic Algorithms, Nearest Neighbour Rule Classifiers (k-NNR).*

## 1. INTRODUCTION

In past works (e.g.: [4],[8],[9]), learning time reduction in Classifiers were achieved by means of training the Classifier with subsets of the initial training set. *Genetic Algorithms* and PCA were then used (*Neural Networks* in the case of [4]) to explore the space of samples, and decide which among them, could maximise the recognition rate.

Among several Classifiers, the *Nearest Neigbhour Rule* is well known [7] and its prove to have successful results even for a small number of training prototypes. However the present paper deals with a different approach: feature reduction.

In that context, a method is described to implement *Nearest Neighbour* optimisation by *Genetic Algorithms* (i.e., via feature reduction). The *Nearest Neighbour rule* (K-NNR) is a simple and powerful classification technique [7]. In the basic *Nearest Neighbour rule* Classifier, each training sample (described by their features) is used as a prototype and a test sample is assigned to the class of the closest prototype [1]. Its asymptotic classification error is bounded above by twice the Bayes error.

When the number of prototypes and/or their feature space is large this method requires a large memory space and long computing time. Because of this, the *Nearest Neighbour rule* has not found wide applications to solve Pattern Recognition problems. For example, the *Nearest Neighbour* Classifier has been shown to have equivalent recognition performance as *Radial Basis Function* (RBF) and *Neural Network* based Classifiers [12]. Although it does not need any training to build a Nearest Neighbour Classifier, it is most expensive to implement it in terms of memory storage and computer time.

Recently Neural Network based methods have been developed to overcome the implementation problem of Nearest Neighbour Classifiers [5]. It has been shown that in some cases a small number of optimised prototypes can even provide a higher classification rate than all training samples used as prototypes.

Another approach consists in using Genetic Algorithms based methods for searching the feature space to apply in Nearest Neighbour Rule prototypes, which is the case in the present paper. For example *Frank Brill et al*

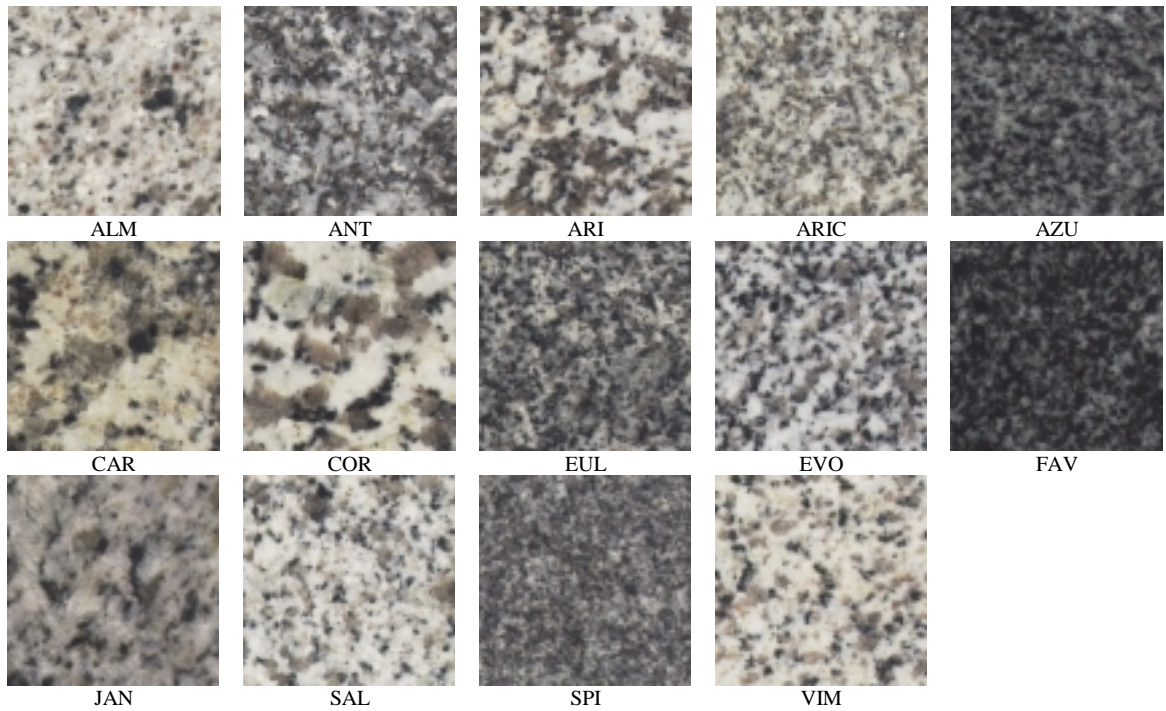

Figure 1 – Portuguese grey granites, commercial designation
(each image represents an area of approximately 4 x 4 cm$^2$)

[3] used a K-NN rule Classifier to evaluate feature sets for counterpropagation networks training. Another authors used the same approach ([10],[11]) for other kinds of Classifiers.

In the present case, a *Nearest Neighbour rule* Classifier was applied to Portuguese Granite Classification. From 237 Granite images, 50 were selected randomly for future performance evaluation (testing set). Each sample on the training set (187 samples) consists on a series of 117 features (HLS histogram values and grey level granulometry data - morphological openings of increasing size - see [6]), which were used on a first NN rule Classifier.

A *Genetic Algorithm* was then built to search the entire space of combinations on these 117 features (exploring the N<117 features), using the classification rate of *the Nearest Neighbour rule* Classifier has his objective function (with Euclidean metric). We describe the Classifier design and implementation procedures and verify the performance of the algorithm with experimental results. Results shown that is possible to use this method, also has a pre-processing tool for other kinds of Classifiers.

## 2. THE TRAINING AND TESTING SET

From the features extracted in each Portuguese Granite image via *Mathematical Morphology* [6], two training/testing sets were build: RGB with 27 features (mainly RGB histogram values) and LOT with 117 features (HLS histogram values and grey level granulometry data - morphological openings of increasing size). On the other hand, the test set was composed with 50 random and independent images representing the 14 different types of Portuguese Granites (see fig.1). For each sample, commercial designations were used - they are:

ALM - Branco Almeida (20 images)
ANT - Cinza Antas (20 images)
ARI - Branco Ariz (8 images)
ARIC - Cinzento Ariz (4 images)
AZU - Azulália (20 images)
CAR - Branco Caravela (20 images)
COR - Branco Coral (20 images)
EUL - Cinzento Stª Eulália (20 images)
EVO - Cinzento Évora (20 images)
FAV - Favaco (15 images)
JAN - Jané (20 images)
SAL - Pedras Salgadas (10 images)
SPI - SPI (20 images)
VIM - Branco Vimieiro (20 images)

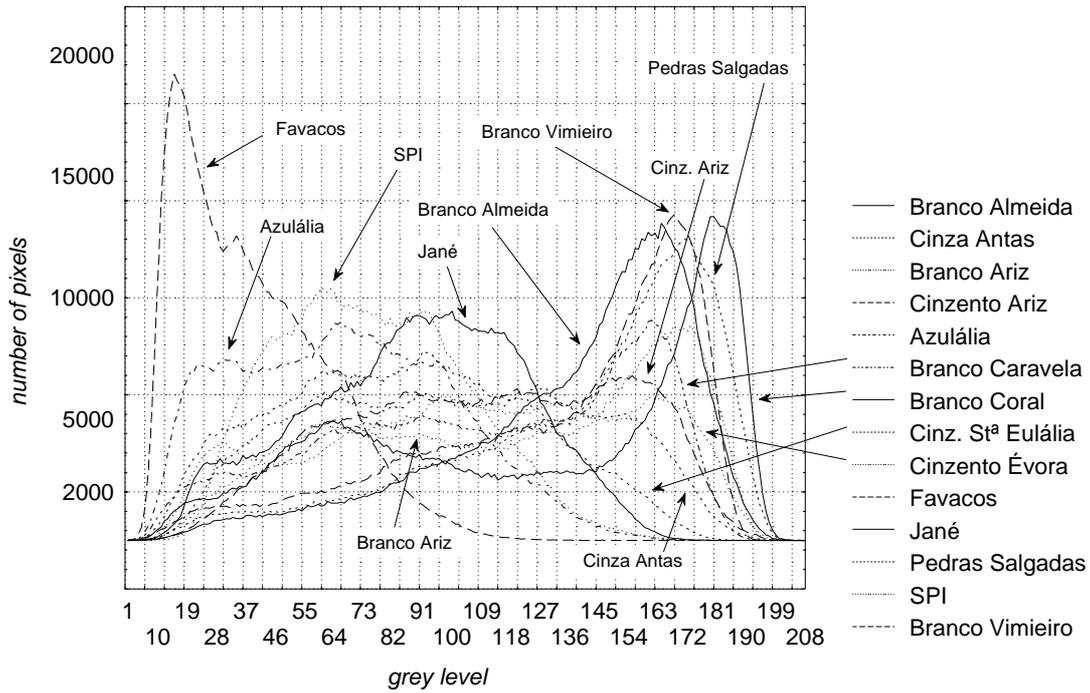

Figure 2 – Intensity Grey level histograms: 14 Portuguese grey granites, commercial designation .

This way, training set RGB is represented by a matrix of 187 lines (187 images representing 14 different types of Portuguese Granites) and 27 columns (27 features). Testing set RGB was represented by a matrix of 50 lines (50 testing images representing also the 14 granites) and 27 columns. Similarly for LOT; i.e., training set with a matrix of 187 lines by 117 columns and the testing set by a matrix of 50 lines and 117 columns.

As said before, feature extraction was mainly done with grey level intensity parameters, for each image: figure 2 represents the grey level Intensity Histograms for the 14 different types of Portuguese granites studied.

## 3. THE *k*-NEAREST NEIGHBOUR RULE

*Nearest Neighbour* methods are among the most popular for classification [7]. They represent the earliest general (nonparametric) methods proposed for this problem and were heavily investigated in the fields of statistics and (especially) pattern recognition.

Recently renewed interest in them has emerged in the connectionist literature ("memory" methods) and also in machine learning ("instance-based" methods). Despite their basic simplicity and the fact that many more sophisticated alternative techniques have been developed since their introduction, nearest neighbour methods still remain among the most successful for many classification problems.

The *K*-Nearest-Neighbour decision rule assigns an object of unknown class to the plurality class among the *K* labelled "training" objects that are close to it. Closeness is usually defined in terms of a metric distance on the Euclidean space with the input measurement variables as axes.

Nearest neighbour methods can easily be expressed mathematically. Let *x* be the feature vector for the unknown input, and let $m_1$, $m_2$, …, $m_c$, be templates (i.e., perfect, noise-free feature vectors) for the *c* classes. Then the error in matching *x* against $m_i$, is given by;

$$\|x - m_i\| \quad (1)$$

Here $\|u\|$ is called the norm of the vector *u*. A minimum-error classifier computes $\|x-m_i\|$ for $i = 1,…c$ and chooses the class for which this error is minimum. Since $\|x-m_i\|$ is also the distance from x to $m_i$, we call this a minimum-distance classifier. Figure 3 shows how the method is implemented. Naturally, the

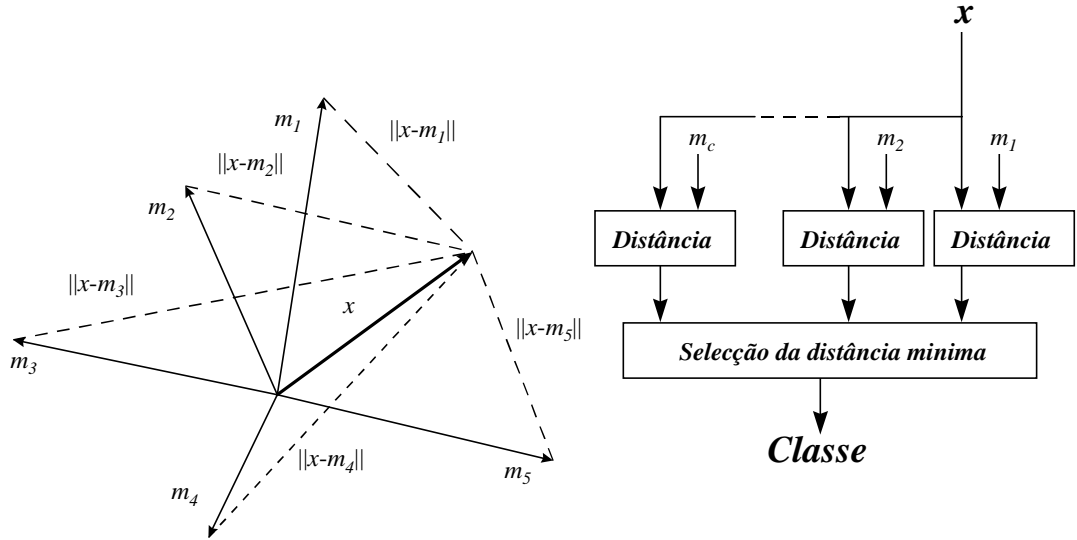

figure 3 - *Nearest Neighbour* Classification. On the left the type of metric, and on the right the classification process for a new sample [9].

$n$ dimension Euclidean distance $d$, of $x$ ($n$ dimensional feature vector) to a training sample $m$, is:

$$d = \sqrt{\sum_{i=1}^{i=N}(x_i - m_i)^2} \qquad (2)$$

The previous approach may be extended to the *k-Nearest Neighbour* rule (k-NNR), where we examine the labels on the *k*-nearest samples in the input space and then classify by using a voting scheme. Often in $c=2$ class problems, $k$ is chosen to be an odd number, to avoid *ties*. Other significant concerns and possible extensions include the use of a reject option in instances where there is no clear *winner*, and the finite sample size performance of the *NNR*.

Given a vector $x$ and a training set $H$, whose cardinality may be large, the computational expense of finding the nearest neighbour of $x$ may be significant. For this reason, frequent attention has been given to efficient algorithms. The computational savings are typically achieved by a preordering of samples in $H$, combined with efficient (often hierarchical) search strategies (for extended analysis see: *I.K.Sethi*; "A Fast Algorithm for Recognising Nearest Neighbours", *IEEE Transactions on Systems, Man and Cybernetics*, Vol. SMC-11, nº3, March, 1981).

## 4. GRANITE CLASSIFICATION

In order to compare future results, 1-*NNR* and 3-*NNR* were implemented using RGB and LOT training sets for Portuguese Granite Classification (i.e., with all features: 27 in RGB; 117 in LOT). Extended results are presented on table 1. The best results in terms of successful recognition were obtained using LOT training set (with 117 features) and 1-NNR (98%). In global terms, recognition errors occur with *Branco Vimieiro* and *Pedras Salgadas* samples.

To get out an idea how these training samples are arranged in their respective $n$ dimensional spaces, *Principal Component Analysis* (PCA) was applied. PCA is perhaps the oldest and best-known technique in multivariate data analysis and data compression (*I.T.Joliffe*; "Principal Component Analysis"; *Springer-Verlag*, 1986). In the present case, PCA has been used to obtain the first two PCs (i.e., the PCs corresponding to the first few dominant eigenvalues of the covariance matrix of the stationary input patterns) and this PCs were used subsequently as 2D *feature* Scatterplot axis (figure 4 shows the scatterplot for the RGB training set; figure 5 for the LOT training set).

## 5. FEATURE SPACE REDUCTION VIA GENETIC ALGORITHMS

In order to reduce the input feature space, and hypothetically improve the recognition rate, Genetic Algorithms were implemented. The idea is to reduce the number of features necessary to obtain at least the same recognition rates. *Genetic algorithms* (GAs -

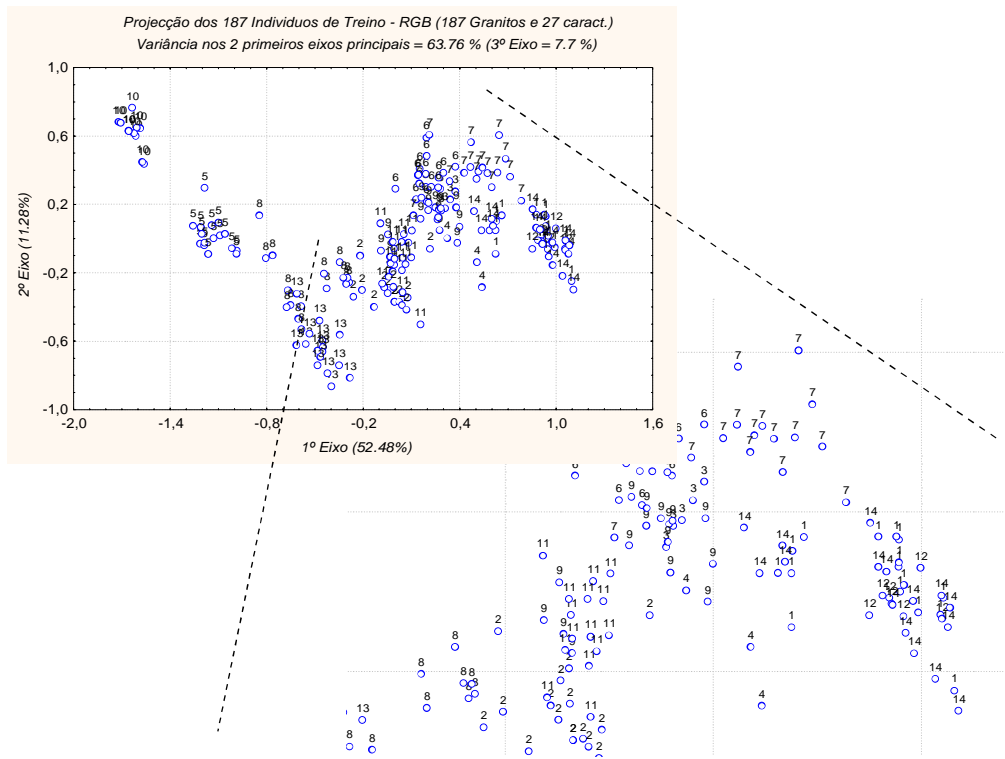

figure 4 - 187 samples - *Principal Component Analysis* projection (187 Portuguese Granite images) from RGB training set (with 27 features for each sample). Each Granite type is refereed: (1) Branco Almeida, (2) Cinza Antas, (3) Branco Ariz, (4) Cinzento Ariz, (5) Azulália, (6) Branco Caravela, (7) Branco Coral, (8) Cinz. St. Eulália, (9) Cinzento Évora, (10) Favaco, (11) Jané, (12) Pedras Salgadas, (13) SPI, e (14) Branco Vimieiro (commercial designation).

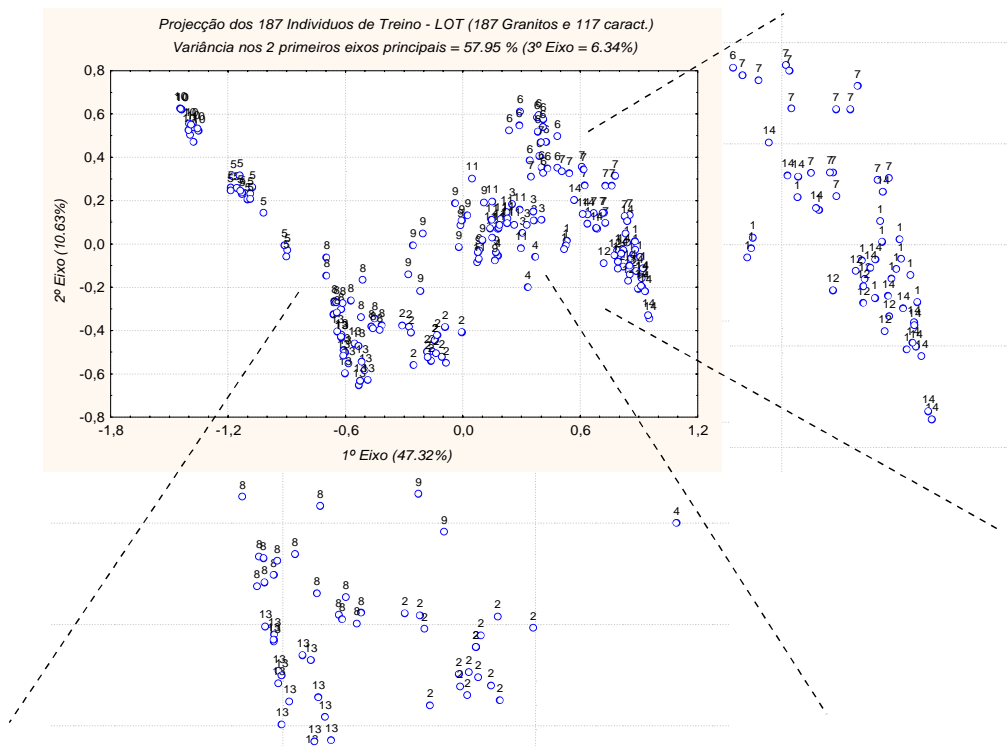

figure 5 - 187 samples - *Principal Component Analysis* projection (187 Portuguese Granite images) from LOT training set (with 117 features for each sample).

|  | *1-NNR RGB* | *1-NNR LOT* | *3-NNR RGB* | | | | *3-NNR LOT* | | | |
|---|---|---|---|---|---|---|---|---|---|---|
| *Recognition Rate* | 96 % | 98 % | 94 % | | | | 98 % | | | |
| *Number of Errors (% in 50)* | 2 = 4 % | 1 = 2 % | 3 = 6 % | | | | 1 = 2 % | | | |
| SAMPLE | Dec. | Dec. | 1ºN. | 2ºN. | 3ºN. | Dec. | 1ºN. | 2ºN. | 3ºN. | Dec. |
| ARI1-2 | **Cor 5-4** | | Cor 5-4 | Vim 2-4 | Vim2-3 | **Vim** | | | | |
| ARIC1-2 | | | Aric 1-4 | Aric 1-3 | Car 4-4 | Aric | | | | |
| EUL1-3 | | | | | | | Eul 1-1 | Eul 1-4 | Ant 2-4 | Eul |
| SAL3-1 | | **Vim 3-1** | Vim 3-1 | Sal 4-1 | Sal 1-1 | Sal | Vim 3-3 | Sal 1-1 | Sal 3-2 | Sal |
| SAL5-1 | **Vim 3-1** | | Vim 3-1 | Vim 4-1 | Vim 3-3 | **Vim** | | | | |
| VIM3-2 | | | | | | | Vim 3-4 | Sal 5-2 | Sal 3-2 | **Sal** |
| VIM4-3 | | | Vim 1-3 | Sal 5-2 | Sal 4-2 | **Sal** | | | | |
| VIM5-1 | | | Vim 4-4 | Vim 5-4 | Sal 2-1 | Vim | | | | |

table 1 - Portuguese Granite Classification - testing Samples for which occurred an recognition error (black cells) or where the voting scheme 3-*NNR* was not in agree. Note the type of *confusions: Branco Ariz* classified as *Branco Coral* (1-*NNR* and 3-*NNR* / RGB), *Pedras Salgadas* as *Branco Vimieiro* (1-*NNR* e 3-*NNR* / RGB ; 1-*NNR* / LOT), and *Branco Vimieiro* as *Pedras Salgadas* (3-*NNR* / RGB and LOT) (Results based on a random and independent testing set with 50 samples).

[2], [9] and [13]) are search procedures based on the mechanics of natural selection and natural genetics. The GA was developed by *John H. Holland* in the 1960s to allow computers to evolve solutions to difficult search and combinatorial problems, such as function optimisation and machine learning. The basic operation of a GA is conceptually simple (canonical GA):

(1) maintain a population of solutions to a problem,
(2) select the better solutions for recombination with each other, and …
(3) use their offspring to replace poorer solutions.

The combination of selection pressure and innovation (through *crossover* and *mutation* - genetic operators) generally leads to improved solutions, often the best found to date by any method (see [2], [13]). For further details on the Genetic operators, GA codification and GA implementation, the reader is also reported to [9].

Usually each individual (*chromosome*; pseudo-solution) in the population (say 50 individuals) is represented by a binary string of 0's and 1's, coding the problem which we aim to resolve. In here the aim is to analyse the combinatorial feature space impact on the Recognition rate of *Nearest Neighbour* Classifiers. In other words, the Genetic Algorithm will explore the $N<117$ features of LOT training set ($N<27$ in the RGB case) and their combinations.

An efficient genetic coding for a feature sub-space is then represented by each GA individual - i.e. an hypothetical classification solution, via a reduced group of features. The GA fitness function is then given by the 1-*NNR* classifier Recognition rate and by the number of features used on that specific *Nearest Neighbour* classifier.

In that way, for each GA generation and for each individual (pseudo-solution) is then performed the Nearest Neighbour classification. The results are then used on the GA again. The algorithm proceeds their search until a stopping criteria is achieved. After convergence, the final solution points out what features among 117 can maximise the fitness function. The overall GA search space is given by the sum of all combinations of 116 features in 117, plus the 115 features in 117,…, plus all combinations of 2 features in 117. For the $i$ individual the fitness function can be expressed as:

$$fit[i] = \alpha hits[i] - \beta nf[i] \quad (3)$$

where $\alpha$ and $\beta$ are real valued constants ($\alpha + \beta = 1$), and $hits[i]$ represents the number of images well recognised among the testing set, and $nf[i]$ the number of features used on *NNR* classification. The representation (GA coding) for each solution is then achieved by means of a binary string 117 bits long (for the LOT training set case; 27 bits long for RGB training set case), i.e. - if the $n^{th}$ bit is 1, then the $n^{th}$ feature is used on the *NNR* classification; if not (= 0) that specific feature is not used on the classification. Similar coding strategies were implemented by several authors, yet for different classification purposes [9].

|  | GA on LOT training set | GA on RGB training set |
|---|---|---|
| nº of original features | 117 | 27 |
| nº of *bits* for each individual | 117 | 27 |
| nº of individuals for each generation | 50 | 216 |
| generations run | 814 | 250 |
| time (hours) | 2.37 | 1.31 |
| Recognition rate on the Test set (50 images) | 100 % | 100 % |
| Crossover probability | 1.00 | 1.00 |
| Mutation probability | 0.90 | 0.90 |
| nº of final features | 3 (de 117) | 5 (de 27) |
| features suggested by the GA - # | 70, 101, 112 | 2, 9, 10, 17, 21 |
| $\alpha$ | 0.6 | 0.4 |
| $\beta$ | 0.6 | 0.4 |
| Random Seed | 12957 | 1547 |

table 2 - GA Parameters used on feature space reduction for classification via *Nearest Neighbour* rule (Computer: PENTIUM 166 MHz / 32 Mb RAM).

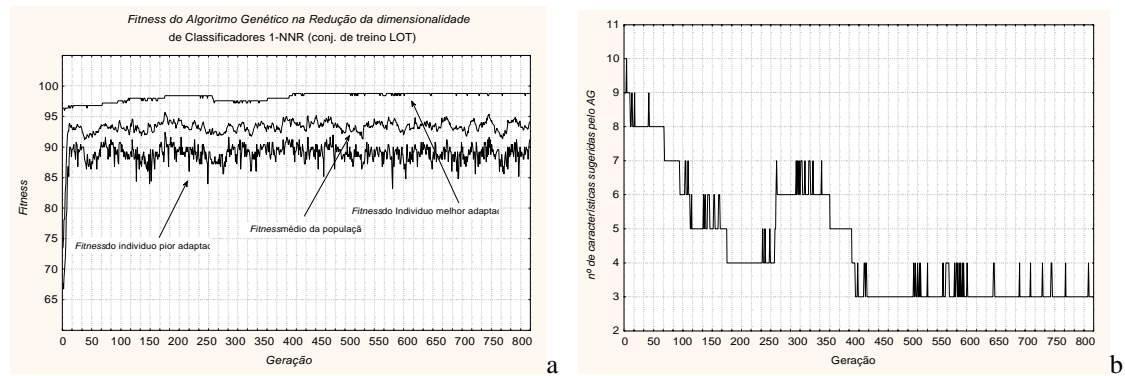

figure 6 - (a) Minimum fitness, best, and median fitness on the 50 GA individual (training set LOT). (b) Number of features in each generation, for the same GA (Portuguese Granite Classification via feature reduction on NNR via Genetic Algorithm).

Table 2 resumes the GA parameters used for feature reduction and classification by means of this hybrid method. Figure 6 (a,b) shows how convergence was achieved, generation after generation, in terms of population fitness (best individual, poor individual, and median population fitness), as well as in terms of feature reduction. Figure 7 gives an idea of the discrimination degree achieved in graphic mode (results for the LOT training set). Since the GA final solution for this set is 3 features, its possible to build 3 scatterplots (each one with 2 features). Finally, figure 7 also presents 3 zooms from one scatterplot, allowing to see in detail each cluster (Granite type).

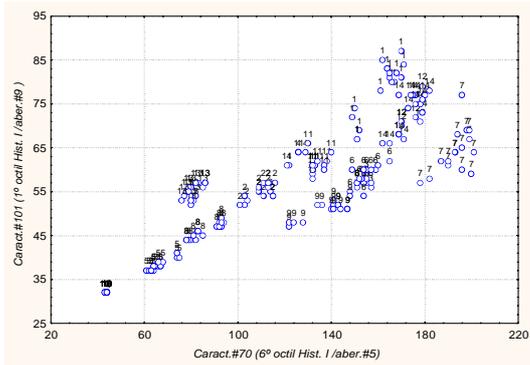 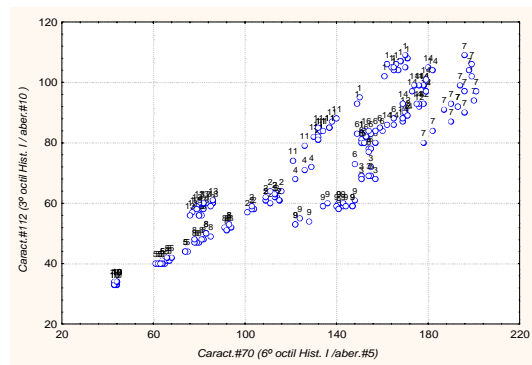

(a)           (b)

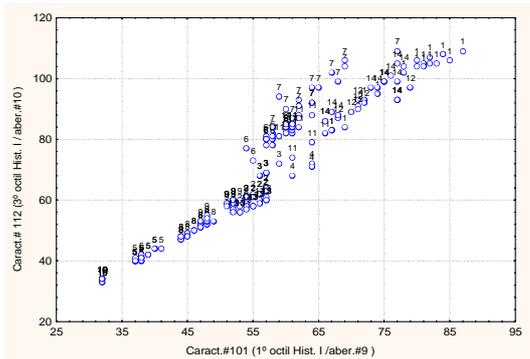 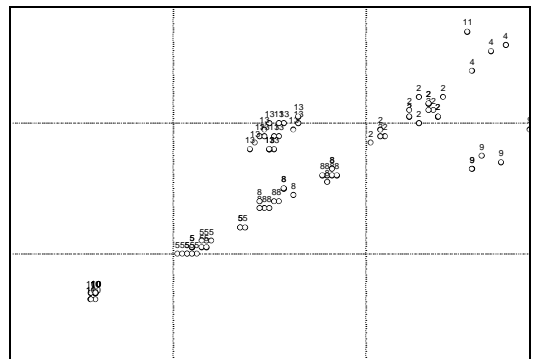

(c)           (d)

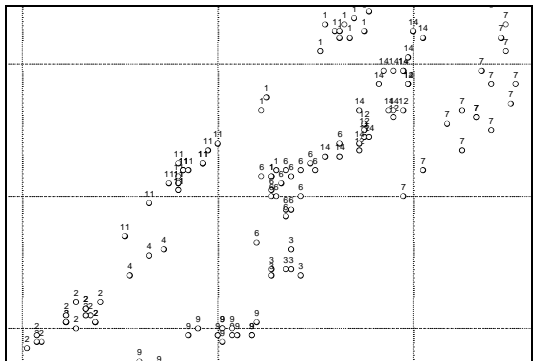 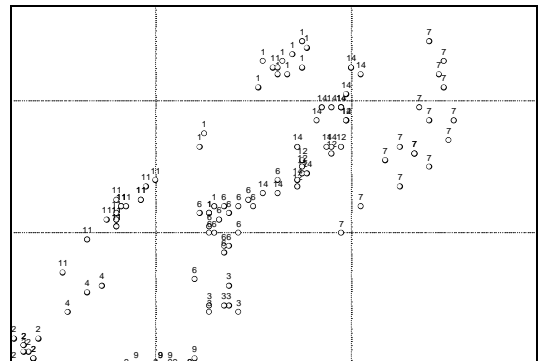

(e)           (f)

figure 7 - 3 scatterplots (a,b,c) resuming the 3D mapping of 187 Portuguese Granite images (LOT training set), via the 3 features (in 117) suggested by the Genetic Algorithm. Sample codes for each type of Granite are the same as in figure 4. 3 Zooms (d,e,f) on the previous scatterplot (figure 7-b) - Features #70 and #112.

## 6. CONCLUSIONS

Results show that this hybrid strategy is highly promising. In fact, not only recognition rate was improved, as the number of features necessary for successful Portuguese Granite image Classification was substantially reduced. Using LOT training set, for instance, was possible to improve the Recognition rate by 2 % (achieving 100%) and reducing the number of important features from 117 to 3 (representing a reduction rate of 97 %).

Computer time was in the last case 142 minutes, but can be reduced using new types of algorithms, strictly involved on the calculus of nearest neighbours (as said before, typically achieved by a preordering of samples in the input space) However, this time is spent only once - i.e., on the training phase. After that, its possible to classify images based on a reduced number of features, improving real time computer calculus on Nearest Neighbour relations, since the dimension of feature space was significantly reduced. On the other hand, this methodology can point out what type of features are really important for successfully discrimination. This way, its possible to understand rigorously and improve knowledge on the proper nature of the digital images we are working.